\documentclass[10pt,twocolumn,letterpaper]{article}

\usepackage{iccv}
\usepackage{times}
\usepackage{epsfig}
\usepackage{graphicx}
\usepackage{amsmath}
\usepackage{amssymb}
\usepackage{multirow}
\usepackage{hhline}
\usepackage{xcolor}

\usepackage[pagebackref=true,breaklinks=true,letterpaper=true,colorlinks,bookmarks=false]{hyperref}
\hypersetup{
    colorlinks = false,
    linkbordercolor = {red},
}

\iccvfinalcopy 

\ificcvfinal \fi

\def\iccvPaperID{381} 

\begin{document}

\title{Bayesian Joint Topic Modelling for Weakly Supervised Object Localisation}

\author{Zhiyuan Shi, Timothy M. Hospedales, Tao Xiang \\
Queen Mary, University of London, London E1 4NS, UK\\
{\tt\small \{zhiyuan.shi,tmh,txiang\}@eecs.qmul.ac.uk}
}
\maketitle
\thispagestyle{empty}

\begin{abstract}
 We address the problem of localisation of objects as bounding boxes in images with weak labels. This weakly supervised object localisation problem has been tackled in the past using discriminative models where each object class is localised independently from other classes. We propose a novel framework based on Bayesian joint topic modelling. Our framework has three distinctive advantages over previous works: (1) All object classes and image backgrounds are modelled jointly together in a single generative model so that ``explaining away'' inference can resolve ambiguity and lead to better learning and localisation.
 (2) The Bayesian formulation of the model enables easy integration of prior knowledge about  object appearance to compensate for limited supervision. (3) Our model can be learned with a mixture of weakly labelled and unlabelled data, allowing the large volume of unlabelled images on the Internet to be exploited for learning. Extensive experiments on the challenging VOC dataset demonstrate that our approach outperforms the state-of-the-art competitors. 
\end{abstract}

\section{Introduction}
Large scale object recognition has received increasing interest in the past five years \cite{DengECCV2010,tim2011tpami,LempitskICCV2009}. Due to the prevalence of online media sharing websites such as Flickr, a lack of images for learning is no longer the barrier. A new bottleneck appears instead: the lack of annotated images, particularly strongly annotated ones. For example, for many vision tasks such as object classification \cite{Nguyenweakly2011}, detection \cite{Felzenszwalb2012partbased}, and segmentation \cite{LempitskICCV2009,Kuettel2012} hundreds or even thousands of object samples must be annotated from images for each object classes. This annotation includes both the presence of objects and their locations, typically in the form of bounding boxes. This is a tedious and time-consuming process that prevents tasks such as object detection from scaling to thousands of classes \cite{Guillaumin_cvpr12}.

\begin{figure}[ht]
\begin{center}
   \includegraphics[width=\linewidth]{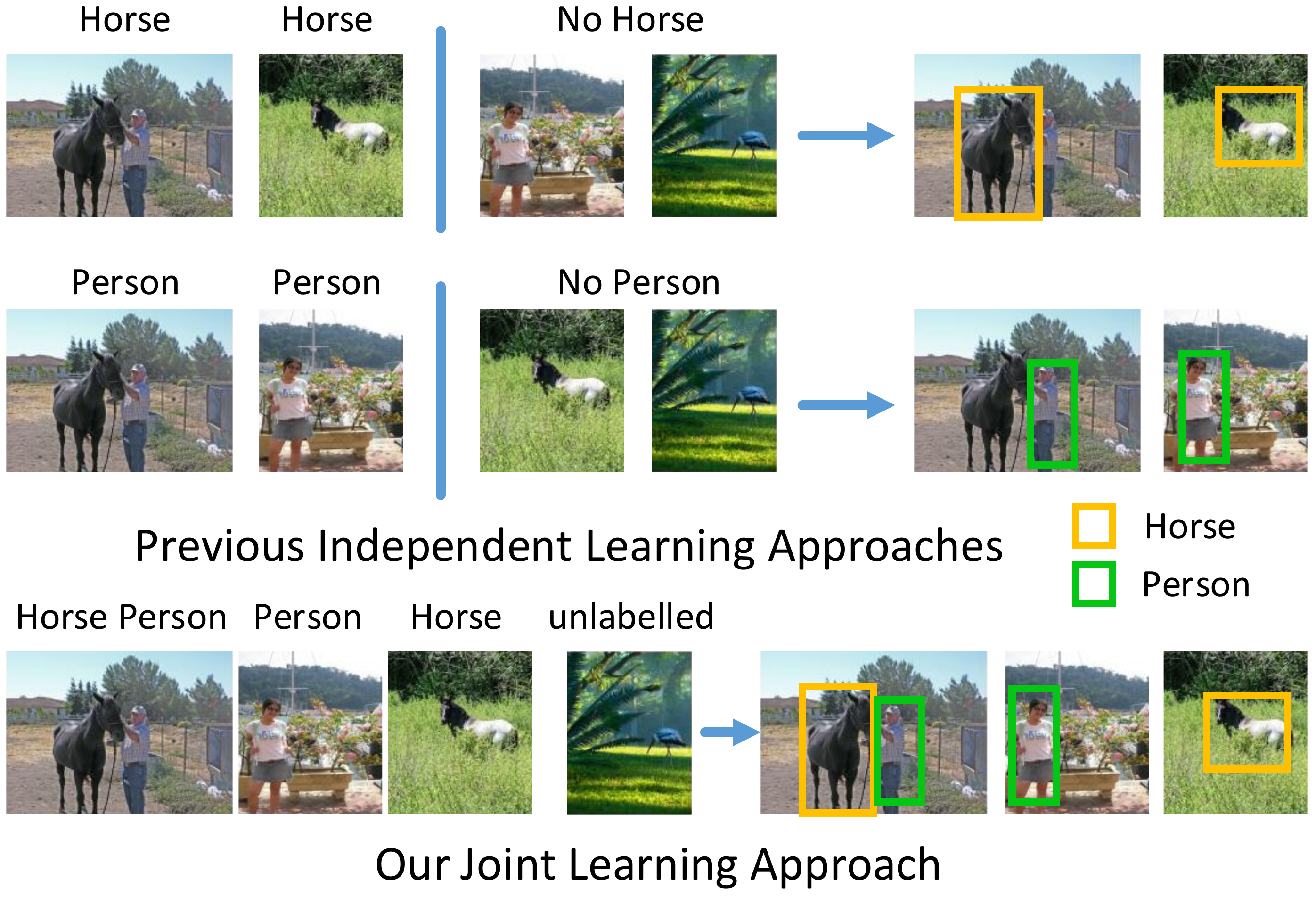}
\end{center}
   \caption{Our jointly learning approach vs. previous approaches which localise each object class independently.}
\label{fig:concept}
\vspace{-10pt}
\end{figure}

One approach to this problem is weakly supervised object localisation (WSOL), which simultaneously locates objects in images and learns their appearance using weak labels indicating only the presence/absence of the object of interest.
The WSOL problem has been tackled using various approaches \cite{Deselaers2012,Nguyenweakly2011,confeccvSivaRX12,Pandeyiccv2011,Guillaumin_cvpr12}. Most of them address the problem as a weakly supervised learning problem, particularly as a multi-instance learning (MIL) problem, where images are bags, and potential object locations are instances. These methods are typically discriminative in nature and attempt to localise each class of objects independently from the other classes, even when the weak labels indicate that different types of objects co-exist in the same images (see Fig.~\ref{fig:concept}). However, localising objects of different classes independently rather than jointly brings about a number of limitations: 
(1) The knowledge that multiple objects co-exist within each image is not exploited. For instance, knowing that some images have both a horse and a person, in conjunction with a joint model for all classes, gives very discriminative information about what a horse and person looks like -- the person can be ``explained away" to reduce ambiguity about the horse appearance, and vice versa. Ignoring this relationship increases ambiguity for each class.
(2) Although different object classes have different appearances, the background appearance is relevant to them all. When different classes are modelled independently, the background appearance must be re-learned repeatedly for each class, when it would be more statistically robust to share this common knowledge between classes.

Beyond joint versus independent learning there is the issue of encoding prior knowledge or top-down cues about appearance, which is very important to obtain good WSOL performance \cite{Deselaers2012,zhiyuan12}. However, the prior knowledge is  typically only employed in existing approaches to provide candidate object locations \cite{Deselaers2012,Sivaiccv2011,confeccvSivaRX12,Pandeyiccv2011,zhiyuan12}, rather than as an integrated part of the model. 
Finally, unlabelled images contain useful information about, e.g. background appearance and the appearance of (the unlabelled) objects. 
Such information is useful when the weak labels are sparse to further reduce the burden for manual annotation. However, existing approaches provide no mechanism for learning from unlabelled data  together with weakly labelled data for object localisation (i.e.~semi-supervised learning (SSL)). This limitation is also related to the lack of joint learning, because for SSL joint learning is important to disambiguate the unlabelled images.

In this paper, a novel framework based on Bayesian latent topic models is proposed to overcome the previously mentioned limitations. In our framework, both multiple object classes and different types of backgrounds are modelled jointly in a single generative model as latent topics, in order to explicitly exploit their correlations (see Fig.~\ref{fig:concept}). As bag-of-words (BoW) models, conventional latent topic models have no notion of localisation. We overcome this problem by incorporating an explicit notion of object location, alongside the ability to incorporate prior knowledge about object appearance in a fully Bayesian approach. Importantly, as a joint generative model, unlabelled data can now be easily used to compensate for sparse training annotations, simply by allowing the model to also infer both which unknown objects are present in those images and where they are.


\section{Related Work}
\label{relatedwork}
\noindent  \textbf{Weakly supervised object localisation} 
Weakly supervised learning (WSL) has attracted increasing attention as the volume of data which we are interested in learning from grows much faster than available annotations. Weakly supervised object localisation (WSOL) is of particular interest \cite{Deselaers2012,Sivaiccv2011,confeccvSivaRX12,Pandeyiccv2011,zhiyuan12,Nguyenweakly2011,Crandalleccv06}, due to the onerous demands of annotating object location information. Many studies \cite{Nguyenweakly2011,Deselaers2012} have approached this task as a multi-instance learning  \cite{Maron98aframework,Andrews03supportvector} problem. However, only relatively recently have localisation models capable of learning from challenging data such as PASCAL VOC 2007 been proposed \cite{Deselaers2012,Sivaiccv2011,confeccvSivaRX12,Pandeyiccv2011,zhiyuan12}. This is especially challenging because objects may occupy only a small proportion of an image, and multiple objects may occur in each image: corresponding to a multi-instance multi-label problem \cite{nguyen2010svm_miml}. One of the first studies to address this was \cite{Deselaers2012} which employed a conditional random field and generic prior object knowledge learned from a fully annotated dataset. Later, \cite{Pandeyiccv2011} presented a solution exploiting latent SVMs. Recent studies have explicitly examined the role of intra- and inter-class cues  \cite{Sivaiccv2011,confeccvSivaRX12}, as well as transfer learning \cite{zhiyuan12,Guillaumin_cvpr12}, for this task. In contrast to these studies, which are all based on discriminative models, we introduce a generative topic model based approach which retains the benefits of both intra- and inter-class cues, as well as the potential for exploiting both spatial and appearance priors. Moreover, it uniquely exploits joint multi-label learning of all object classes simultaneously, as well as enables semi-supervised learning which allows annotation requirements to be further reduced.

\noindent  \textbf{Topic models for image understanding}  Topic models were originally developed for unsupervised text analysis \cite{BleiLDA2003}, and have been successfully adapted to both unsupervised \cite{Philbinijcv2010,Sivic05b} and supervised image understanding problems \cite{CaoFei-Fei2007,LiSocherFeiFei2009,wangbleifeifei08}. Most studies have addressed the simpler tasks of learning classification \cite{wangbleifeifei08,CaoFei-Fei2007,LiSocherFeiFei2009} or annotation \cite{wangbleifeifei08,LiSocherFeiFei2009,blei2003annotated_model}, rather than localisation which we are interested in here. This is because conventional topic models have no explicit notion of the spatial location and extent of an object in an image; and because supervised topic models such as CorrLDA \cite{blei2003annotated_model} and derivatives \cite{wangbleifeifei08} allow much less direct supervision than we will exploit here.
Most of these studies have considered smaller scale and simpler datasets than  VOC 2007, which we consider here. Nevertheless topic models have good potential for this challenge because they can  be modified for multi-label weakly supervised learning \cite{fu2012attribsocial, tim2011tpami}, and can then reason jointly about multiple objects in each image. Moreover as generative models, they can  be easily applied in a semi-supervised learning context \cite{zhu2007sslsurvey} and Bayesian versions can exploit informative priors \cite{feifei2006one_shot}. In this paper we address the limitations of existing topic models for this task by incorporating an explicit notion of object location; and developing a Bayesian model with the ability to incorporate prior knowledge about object appearance (e.g.~texture, size, spatial extent).

\noindent \textbf{Other joint learning approaches}  An approach similar in spirit to ours in the sense of jointly learning a model for all classes is that of Cabral \etal \cite{CabralDCB11}. This study formulates multi-label image classification as a matrix completion problem, which is also similar in spirit to our factoring images into a mixture of topics. However we add two key factors of (i) a stronger notion of the spatial location and extent of each object, and (ii) the ability to encode human knowledge or transferred knowledge through Bayesian priors. As a result we are able to address more challenging data than \cite{CabralDCB11} such as VOC 2007.
Multi-instance multi-label (MIML) \cite{nguyen2010svm_miml} approaches provide a mechanism to jointly learn a model for all classes \cite{Zhou07multi-instancemultilabel,zha2008mlmi}. However because these methods must search a discrete space (of positive instance subsets), their optimisation problem is harder. They also lack the benefit of Bayesian integration of prior knowledge.
Finally, while there exist more elaborative joint generative learning methods \cite{sudderth2008tdp_visual,LiSocherFeiFei2009}, they are more complicated than necessary for the WSOL task and thus do not scale to the size of data required here.

\noindent \textbf{Our contributions} are threefold: (1) We propose the novel concept of joint modelling of all object classes and backgrounds for weakly supervised object localisation. (2) We formulate a novel Bayesian topic model suitable for localisation of object and utilising various types of prior knowledge available. (3) We provide a solution for exploiting unlabelled data for weakly supervised object localisation. Extensive experiments show that our model  surpasses all existing competitors and achieves state-of-the-art performance.

\section{Methods}

In this section, we introduce our new  latent topic model (LTM) \cite{BleiLDA2003} approach to the weakly-supervised
object localisation task, and the associated learning algorithms. Applied to
images, conventional LTMs factor images into combinations of latent
topics \cite{Philbinijcv2010,Sivic05b}. Without supervision, these
topics may or may not correspond to anything of semantic relevance
to humans. To address the WSOL task, we need to learn what is unique
to all images sharing a particular label (object class), explaining away other shared
visual aspects (background) which are irrelevant to the annotation of interest.
We will achieve this in a fully Bayesian LTM framework by applying weak supervision
to partially constrain the available topics for each image.

\subsection{Preprocessing and Representation}

We preprocess images by extracting $N_{j}$ SIFT descriptors, regularly sampled every 5 pixels, and quantising them into a $N_{v}=2000$
word codebook using K-means clustering. Differently to other bag-of-words (BoW)
approaches \cite{LiSocherFeiFei2009,wangbleifeifei08}
which then discard spatial information entirely, we then represent
each image $j$ by a list of $N_j$ words and corresponding locations
$\{x_{i},l_{xi},l_{yi}\}_{i=1}^{N_j}$. Although we solely use SIFT here, other 
BoW type features can easily be included to further increase performance.

\subsection{Our Framework}

\begin{figure}[t]
\begin{center}
   \includegraphics[width=0.8\linewidth]{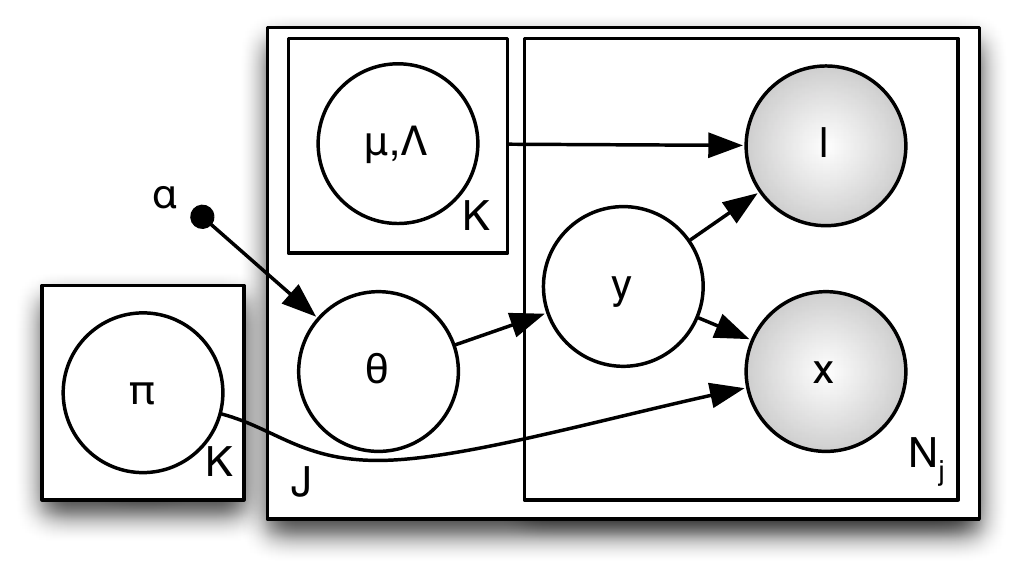}
\end{center}
   \caption{Graphical model for WSOL Joint topic Model. Shaded nodes are observed.}
\label{fig:GM}
\end{figure}

\paragraph{Model}

To address the WSOL task, we will factor images into unique combinations
of $K$ shared topics. If there are $C$ classes of objects to
be localised, $K^{fg}=C$ of these will represent the (foreground) classes,
and $K^{bg}=K-K^{fg}$ topics will model background
data to be explained away. $T^{fg}$ and $T^{bg}$ index foreground
and background topics respectively. Each topic will encode a distribution
over the $N_{v}$ sized appearance vocabulary, and over the spatial
location of these words within each image. Formally, the generative
process of our model (Fig.~\ref{fig:GM}) for a corpus of images is as follows: 

\noindent For each topic $k\in1\dots K$:
\begin{enumerate}
\item Draw an appearance distribution $\pi_{k}\sim\mbox{Dir}(\pi_{k}^{0})$
\end{enumerate}
For each image $j\in1\dots J$:
\begin{enumerate}
\item Draw foreground and background topic distribution $\theta_{j}\sim\mbox{Dir}(\alpha_{j})$,
$\alpha_{j}=[\alpha_{j}^{fg},\alpha_{j}^{bg}]$.
\item For each foreground topic $k\in T^{fg}$ draw a location distribution:
$\{\mu_{kj},\Lambda_{kj}\}\sim\mathcal{NW}(\mu_{k}^{0},\Lambda_{k}^{0},\beta_{k}^{0},v_{k}^{0})$ 
\item For each observation $i\in1\dots N_j$:

\begin{enumerate}
\item Draw topic $y_{ij}\sim\mbox{Multi}(\theta_{j})$
\item Draw visual word $x_{ij}\sim\mbox{Multi}(\pi_{y_{ij}})$
\item Draw a location: $\mathbf{l}_{ij}\sim\mathcal{N}(\mu_{y_{ij}j},\Lambda_{y_{ij}j}^{-1})$
if $y_{ij}\in T^{fg}$; or $\mathbf{l}_{ij}\sim Uniform$ if $y_{ij}\in T^{bg}$
\end{enumerate}
\end{enumerate}

\noindent where Multi, Dir, $\mathcal{N}$ and $\mathcal{NW}$ respectively indicate Multinomial, Dirichlet, Normal and Normal-Wishart distributions with the specified parameters. These priors are chosen because they are  conjugate to the word and location distributions, and hence enable efficient inference. The joint distribution of all observed $O=\{\mathbf{x}_{j},\mathbf{l}_{j}\}_{j=1}^{J}$
and latent $H=\{\{\pi_{k}\}_{k=1}^{K},\{\mathbf{y}_{j},\mu_{kj},\Lambda_{kj},\theta_{j}\}_{k=1,j=1}^{K,J}\}$
variables given parameters $\Pi=\{\{\pi_{k}^{0},\mu_{k}^{0},\Lambda_{k}^{0},\beta_{k}^{0},v_{k}^{0}\}_{k=1}^{K},\{\alpha_{j}\}_{j=1}^{J}\}$
in our model is therefore:
\begin{eqnarray}
p(O,H|\Pi) & = & \prod_{j}^{J}\prod_{k}^{K}\left[\vphantom{\prod_{i}^{N_{i}}}p(\mu_{jk},\Lambda_{jk}|\mu_{k}^{0},\Lambda_{k}^{0},\beta_{k}^{0},v_{k}^{0})\right.\nonumber \\
 &  & \left.\hspace{-1.5cm}\hspace{-1cm}p(\theta_{j}|\alpha_{j})\left(\prod_{i}^{N_j}p(x_{ij}|y_{ij},\theta_{j})p(y_{ij}|\theta_{j})\right)\right]p(\pi_{k}|\pi_{k}^{0}).\label{eq:Joint}
\end{eqnarray}
\paragraph{Learning}

Learning our model involves inferring the following quantities: the
appearance of each object class, $\pi_{k},k\in T^{fg}$ and background
textures, $\pi_{k},k\in T^{bg}$, the word-topic distribution (soft segmentation) of each image $\mathbf{z}_{j}$,
the proportion of interest points in each image corresponding to each
class or background $\theta_{j}$, and the location of each object
$\mu_{jk},\Lambda_{jk}$. To learn the model and localise all the weakly
annotated objects, we wish to infer the posterior $p(H|O,\Pi)=p(\{\mathbf{y}_{j},\mu_{jk},\Lambda_{jk},\theta_{j}\}_{k,j}^{K,J},\{\pi_{k}\}_{k}^{K}|\{\mathbf{x}_{j},\mathbf{l}_{j}\}_{j=1}^{J},\Pi)$.
This is directly intractable, however a variational message passing
(VMP) \cite{winn2004vmp} strategy can be used to obtain a factored
approximation $q(H|O,\Pi)$ to the posterior:
\begin{eqnarray}
q(H|O,\Pi) & =\nonumber \\
 &  & \hspace{-1.5cm}\prod_{k}q(\pi_{k})\prod_{j}q(\theta_{j})q(\mu_{jk},\Lambda_{jk})\prod_{i}q(y_{ij}).\label{eq:varApprox}
\end{eqnarray}
\noindent The VMP solution is obtained by deriving integrals of the
form $\ln q(\mathbf{h})=E_{H\backslash\mathbf{h}}\left[\ln p(H,O)\right]+K$
for each group of hidden variables $\mathbf{h}$, thus obtaining the
updates: 
\begin{eqnarray}
\theta_{jk} & = & \alpha_{jk}+\sum_{i} y_{ijk}, \nonumber \\
y_{ijk} & \propto & \int_{\mu_{jk},\Lambda_{jk}}\mathcal{N}(\mathbf{l}_{ij}|\mu_{jk},\Lambda_{jk}^{-1})q(\mu_{jk},\Lambda_{jk}) \nonumber \\
 &  & \hspace{-1.5cm}\cdot\exp\left(\Psi(\pi_{x_{ij}y_{ij}})-\Psi(\sum_{x}\pi_{xy_{ij}})+\Psi(\theta_{jy_{ijk}})\right), \nonumber \\
\pi_{vk} & = & \pi^0_{vk}+\sum_{ij}\mathbf{I}(x_{ij}=v) y_{ijk}, \label{eq:varUpdates}
\end{eqnarray}
\noindent where $\Psi$ is the digamma function, $\mathbf{I}$ is
the indicator function which returns 1 if its argument is true, and
the integral in the second line returns a student-t distribution over
$\mathbf{l}_{ij}$. Within each image $j$, standard updates apply
for the Gaussian parameter posterior $q(\mu_{jk},\Lambda_{jk})$ \cite{bishop2006prml},
which we omit for brevity. 

In conventional topic models the $\alpha$ parameter encodes the expected proportion of words for each topic. In this study we use $\alpha$ to encode the supervision from weak labels. In particular, we set $\alpha_{j}^{fg}$
as a binary vector with $\alpha_{jk}^{fg}=1$ if class $k$ is present
in image $j$ and $\alpha_{jk}^{fg}=0$ otherwise; $\alpha^{bg}$
is always set to $1$ to reflect the fact that background of different types can be shared across different images. With these partial constraints, iterating the
updates in Eq.~(\ref{eq:varUpdates}) has the effect of factoring images
into combinations of latent topics; where $K^{bg}$ background
topics are always available to explain away backgrounds, and
$K^{fg}$ foreground topics are only available to images with
annotated classes. After learning, we can also localise in held-out
test data by fixing $q(\pi_{vk})$, and iterating the other updates
for test images.

\noindent \textbf{Encoding human or transferred knowledge via Bayesian prior}
An important capability of our Bayesian approach is that top-down prior knowledge
from human expertise, or other transferrable cues can be encoded. A number of different types of human knowledge about objects and their relationships with backgrounds are encoded in our model. First, objects are typically compact whilst backgrounds much less so and tend to spread across the image. This knowledge is encoded via the Gaussian foreground topic spatial distribution
and the uniform background topic distribution. (see 3.(c) in generative process) Second, aggregated across all images, the background is more dominant than any single object class in terms of size (hence the amount of visual words). Consequently, for each object class $k$, we set $\pi^0_k = \left|\frac{1}{N_c}\sum_{j,c_j=k} h(\mathbf{x}_j)-\frac{1}{N_j}\sum_j h(\mathbf{x}_j) \right|_+$, where $h(\cdot)$ indicates histogram. That is, set the appearance prior for each class to the mean of those images containing the object class minus the average over all images, which reflects consistent unique aspects of that class. This prior knowledge is essentially the fact that foreground objects stand out against background, and thus is related to the notion of saliency, not within an image, but across all images. Saliency has been exploited in previous MIL based approaches to generate the instances/candidate object locations \cite{Deselaers2012,Sivaiccv2011,confeccvSivaRX12,Pandeyiccv2011,zhiyuan12}. Here in our model, it is fully integrated as Bayesian prior. Apart from these two types of human knowledge, other human or transferrable knowledge extracted from auxiliary labelled data can also be readily integrated into our model via the Bayesian priors. 
For example, if 
there is prior knowledge about the appearance of individual classes
(e.g., by obtaining the opinion of a generic object detector or object saliency model \cite{Alexewhatisobject}
on images labelled with class $c$), then this can be encoded
via the appearance prior by specifying an informative $\pi_{c}^{0}$
set to the average statistics of the generic object bounding-boxes. In summary, our Bayesian joint topic model is flexible and versatile in allowing use of any  knowledge available additional to the weak labels. \\
\noindent \textbf{Semi-supervised learning }
Our framework can be applied in a semi-supervised
context to further reduce the amount of annotation effort required.
Specifically, images $j$ with known annotations are encoded as above, while those of
unknown class are simply set as $\alpha_{j}^{fg}=0.1$, meaning that
all topics/classes may occur, but we expect few
at once within one image. Importantly, unknown images can include those from
the same pool of classes but without annotation (for which the posterior
$q(\theta)$ will pick out the present classes), or those from a completely
disjoint pool of classes (for which the $q(\theta)$ will
encode only background).

\subsection{Object Localisation\label{sub:Object-Localization}}

There are two possible strategies to localise objects in our framework,
which we will compare later. In the first strategy (\textit{Our-Gaussian}), a bounding box
for class $k$ in image $j$ can be obtained directly from the Gaussian
mode of $q(\mu_{jk},\Lambda_{jk})$,
via aligning a window to the two standard deviation ellipse.  This has the advantage of being
clean and highly efficient. However, since there
is only one Gaussian per class (which will grow to cover all instances
of the class in an image), this is not ideal for images with more
than one object per class. In the second strategy (\textit{Our-Sampling}) we draw a heat-map for class $k$ by projecting $q(z_{ijk})$ (Eq.~(\ref{eq:varUpdates})) back onto the image plane,
using the SIFT grid coordinates. This heat-map is analogous to those
produced by many other approaches such as Hough transforms \cite{houghforest2012}.
Thereafter, any strategy for heat-map based localisation may be used.
We choose the non-maximum suppression (NMS) strategy of \cite{Felzenszwalb2012partbased}.

\section{Experiments}
After briefly introducing the evaluation datasets, we first compare our model with state-of-the-art on localising objects in weakly annotated images in Sec.~\ref{soa}. In Sec.~\ref{semi-supe} we demonstrate that our model is able to effectively exploit semi-supervised learning to further reduce annotation requirements. Then we give the insight into the mechanisms and novelties of our model in Sec.~\ref{modellearned} by illustrating the learned internal representation and comparing against alternative learning methods. 
Finally in Sec.~\ref{cost}, we discuss computational efficiency.
 
\noindent \textbf{Datasets}\quad We use the challenging PASCAL VOC 2007 dataset that has become widely used for weakly supervised annotation. A number of variants are used: \textit{VOC07-20} contains all 20 classes from VOC 2007 training set as defined in \cite{Sivaiccv2011} and has been used in \cite{Sivaiccv2011,confeccvSivaRX12,zhiyuan12}; \textit{VOC07-6$\times$2} contains 6 classes with Left and Right poses considered as separate classes giving 12 classes in total and has been used in \cite{Deselaers2012,Pandeyiccv2011,Sivaiccv2011,confeccvSivaRX12,zhiyuan12}.
The former obviously is more challenging than the latter. Note that \textit{VOC07-20} is different to the \textit{Pascal07-all} defined in  \cite{Deselaers2012} which actually contains 14 classes as the other 6 were used as fully annotated auxiliary data. We call it \textit{VOC07-14} for evaluation against \cite{Deselaers2012,Pandeyiccv2011,bMCL2012}, but without using the other 6 as auxiliary data for our model. \\
\noindent \textbf{Settings}\quad For our model, we set the foreground topic number $N^{fg}$ to be equal to the number of classes, and $N^{bg}=20$ for background topics. We run Eq.~(\ref{eq:varUpdates}) for $100$ VMP iterations. Localisation performance is measured according to the PASCAL criterion \cite{pascalvoc2007}: the object is considered as correctly localised if the overlap with ground-truth is greater than 50\%, and results are reported as the percentage (\%) of correctly annotated images.\\

\subsection{Comparison with State-of-the-art}
\label{soa}
\begin{table}[ht]
\scriptsize
\begin{center}
\begin{tabular}{l | l | l| l |l | l| l}
\hline
\multicolumn{1}{c|}{\multirow{2}{*}{Method} }&\multicolumn{3}{c|}{Initialisation} & \multicolumn{3}{c}{Refined by detector} \\
\hhline{~------}

&  \textit{6$\times$2} & \  \textit{14} & \ \textit{20} & \textit{6$\times$2}  & \  \textit{14}& \ \textit{20} \\

\hline
\hline
Deselares \etal  \cite{Deselaers2012} \\
\hhline{~------}
\hspace{10pt}    a. single cues & 35 & 21  & - & 40  & 24  & - \\
\hhline{~------}
\hspace{10pt}    b. all cues & 39 &  22   & -  & 50  &  28   & - \\
\hline
Pandey and Lazebnik \cite{Pandeyiccv2011} $^*$   \\
\hhline{~------}
\hspace{10pt}    a. before cropping & 36.7 & 20.0 & -  & 59.3& 29.0  & - \\
\hhline{~------}
\hspace{10pt}    b. after cropping & 43.7&  23.0 & -  & 61.1 & 30.3 & - \\
\hline
Siva and Xiang \cite{Sivaiccv2011} & 40  & -  & 28.9 & 49  & - & 30.4 \\
\hline
Siva \etal \cite{confeccvSivaRX12} & 37.1& -  & 29.0  & 46   & - \\
\hline
Shi \etal \cite{zhiyuan12} $^+$   & 39.7 & -  & 32.1 & - & -  & - \\
\hline
Zhu \etal \cite{bMCL2012}  & - & -  & - & - & 31   & - \\
\hline
\hline
Our-Sampling  &  50.8   & \textbf{32.2} &\textbf{34.1}&  65.5 & \textbf{33.8} &  \textbf{36.2} \\
\hline
Our-Gaussian  & \textbf{51.5}  & 30.5  &  31.2  &  \textbf{66.1} & 32.5 &  33.4 \\
\hline

\end{tabular}
\end{center}
\caption{Comparison with state-of-the-art competitors on the three variations of the PASCAL VOC 2007 datasets. \footnotesize{$^*$  Requires aspect ratio to be set manually. $^+$ Require 10 out of the 20 classes fully annotated with bounding-boxes and used as auxiliary data.}} 
\label{state-of-art}
\end{table}

\noindent \textbf{State-of-the-art competitors } Most if not all recent approaches that report results on at least one of the three variants of the VOC 2007 datasets are listed in Table \ref{state-of-art}. These cover a variety of approaches which use very different cues and models (see Sec.~\ref{relatedwork} for details). Their performance is compared against two variations of our models, \textit{Our-Sampling} and \textit{Our-Gaussian} which differ only in the final object localisation step (see Sec.~\ref{sub:Object-Localization}). Note that a number of the state-of-the-art competitors use additional information that we do not use:  Deselares et al.~\cite{Deselaers2012} and Shi et al.~\cite{zhiyuan12} take a transfer learning approach and require a fully annotated auxiliary dataset. In particular,  although Shi et al.~\cite{zhiyuan12} evaluate all 20 classes, a randomly selected 10 are used as auxiliary data with bounding-boxes annotations. Pandey and Lazebnik \cite{Pandeyiccv2011} set aspect ratio manually and/or performs cropping on the obtained bounding-boxes.

\noindent \textbf{Initial localisation } Table \ref{state-of-art} reports the initial annotation accuracy of our model compared with state-of-the-art. Our model shows superior performance on all datasets.  This is because we uniquely provide a jointly-multi label model, and also can exploit prior spatial and appearance cues in an integrated Bayesian framework. 



\noindent \textbf{Refined by detector} After the initial annotation of the weakly labelled images, a conventional strong object detector can be trained using these annotations as ground truth. The trained detector can then be used to iteratively refine the object location.   We follow \cite{Pandeyiccv2011,Sivaiccv2011} in exploiting a deformable part-based model (DPM) detector  \cite{Felzenszwalb2012partbased} for one iteration to refine the initial annotation. Table \ref{state-of-art} shows that again our model outperforms all competitors by a clear margin for all three datasets. In particular, even after this costly refinement process, the localisation accuracy of many competitors is inferior to our model without the refinement. The results also show that the improvement brought by the refinement can be very limited or even negative for some classes when the initialisation performance is poor (see supplementary material for more detailed comparisons).


\subsection{Semi-supervised Learning} 
\label{semi-supe}
One important advantage of our model is the ability to utilise completely unlabelled data, to further reduce the manual annotation requirements. To demonstrate this we randomly select $10\%$ of the \textit{VOC07-6$\times$2} data as weakly labelled training data, and then add different unlabelled data. Note that 10\% labelled data corresponds to around only \textit{5 weakly labelled images per class} for the  \textit{VOC07-6$\times$2} dataset, which is significantly less than any previous method exploits. Varying the unlabelled data used, the following conditions are considered: (i) Only the 10\% labelled data are included (\textit{10\%L}); (ii) The remaining ($90\%$) of $6\times2$ data (\textit{10\%L+90\%U}) are included without any annotation (so the unlabelled data contains relevant but un-differentiated classes); (iii) The remaining VOC07 training data (\textit{10\%L+AllU}) are included ($90\%$ of the \textit{6$\times$2} data and $100\%$ of the remaining 14 classes). This is the most realistic scenario, reflecting including an easy-to-obtain pool of data containing both related and un-related images. Finally, two evaluation procedures are considered: (i) Evaluating localisation performance on the initially annotated $10\%$ (standard WSOL task); and (ii) WSOL performance on the held out  \textit{VOC07-6$\times$2} test set. This latter procedure corresponds to an online application scenario where the localisation model is trained on one database and needs to be applied online to localise objects in incoming weakly labelled images.

\begin{table}[ht]
\footnotesize
\begin{center}
\begin{tabular}{l || c | c  }
\hline
VOC07-$6\times2$ & \multicolumn{2}{c}{Data for Localisation}\\
\hline
Data for Training &  \textit{10\%L} & \textit{Test set} \\

\hline
\hline
\textit{10\%L}  & 27.1   &   28.0  \\
\hline
\textit{10\%L+90\%U}  & 47.1  &   42.3   \\
\hline
\textit{10\%L+AllU}  & 46.8   &   43.8   \\
\hline
\textit{100\%L}  & 50.3   &   46.2  \\
\hline
\end{tabular}
\end{center}
\caption{Semi-supervised learning performance of \textit{Our-Sampling}.}
\label{tab:sslBar}
\end{table}

\begin{figure}[t]
\begin{center}
   \includegraphics[width=\linewidth]{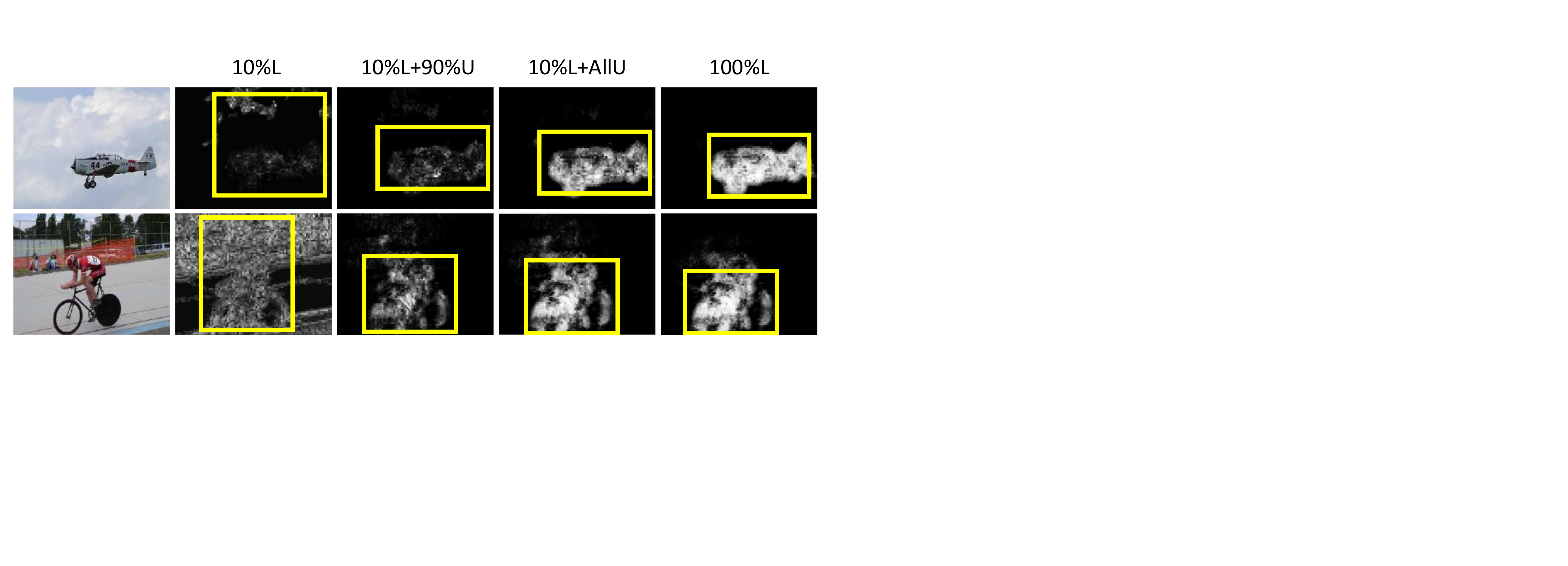}
\end{center}
   \caption{Unlabeled data improves foreground heat maps.}
\label{fig:semilearned}
\end{figure}

From the results shown in Table \ref{tab:sslBar}, our model is clearly capable of exploiting unlabelled data to good effect. With only 5 images per class, as expected, poor results are obtained (comparing \textit{10\%L} with \textit{100\%L}). However if the remaining $90\%$ of the data can be provided \emph{unannotated}, performance is only a few percent below the fully annotated case (comparing \textit{10\%L+90\%U} with \textit{100\%L}). More impressively, even if only a third of the provided unlabelled data is at all relevant (\textit{10\%L+AllU}), good performance is still obtained. This result shows that our approach has good promise for effective use in economically realistic scenarios of learning from only few weak annotations and a large volume of only partially relevant unlabelled data. 
This is illustrated visually in Fig.~\ref{fig:semilearned}, where unlabelled data clearly helps to learn a better object model.
Finally, the similarly good results on the held-out test set verify that our model is indeed learning a good generalisable localisation mechanism and is not merely over fitting to the training data. 

\subsection{Insights into Our Model} 
\label{modellearned}
\begin{figure*}[t]
\begin{center}
   \includegraphics[width=\linewidth]{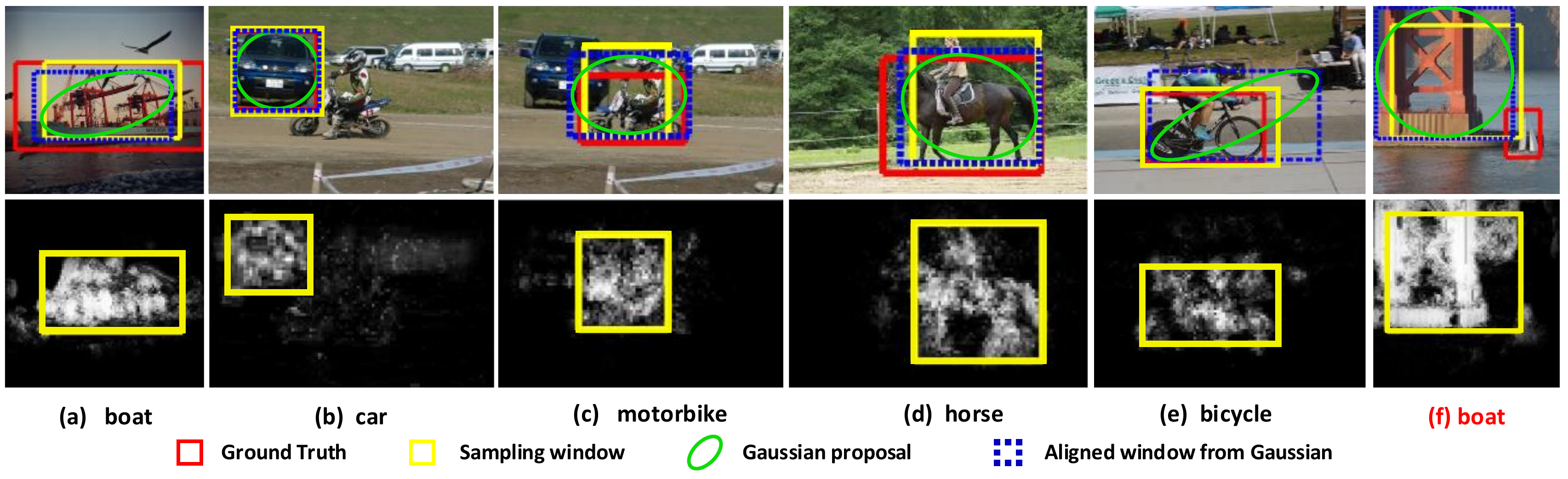}
\end{center}
   \caption{Illustration of the object localisation process and what are learned by the object (foreground) topics using the heat map in the bottom row (higher intensity values mean higher model response). 
   }
\label{fig:objecttopic}
\end{figure*}

\noindent \textbf{Object localisation and learned foreground topics } Qualitative results are illustrated in Fig.~\ref{fig:objecttopic}, including heat maps of the object location showing what has been learned by those object (foreground) topics in our model. The predicted Gaussian object locations (green and blue) are shown along with those obtained by sampling the heat maps. These examples show that the foreground topics indeed capture what each object class looks like and can distinguish it from background and between different object classes. For instance, Fig.~\ref{fig:objecttopic}\textcolor{red}{(b)} and \ref{fig:objecttopic}\textcolor{red}{(c)} illustrate the object of interest ``explain away'' other objects of no interest. A car is successfully located in Fig.~\ref{fig:objecttopic}\textcolor{red}{(b)} using the heat map of the car topic, while Fig.~\ref{fig:objecttopic}\textcolor{red}{(c)} shows that the motorbike heat map is quite accurately selective, with minimal response obtained on the other vehicular clutter. Fig.~\ref{fig:objecttopic}\textcolor{red}{(d)} indicates how the Gaussian  can sometimes give a better bounding box. The opposite is observed in Fig.~\ref{fig:objecttopic}\textcolor{red}{(e)} where the single Gaussian assumption is not ideal when the foreground topic has less a compact response. Finally a failure case is shown in Fig.~\ref{fig:objecttopic}\textcolor{red}{(f)}, where a bridge structure resembles the boat in Fig.~\ref{fig:objecttopic}\textcolor{red}{(a)} resulting strong response from the foreground topic, whilst the actual boat, although picked up by the learned boat topic, is small and overwhelmed by the false response.

\begin{figure}[t]
\begin{center}
   \includegraphics[width=\linewidth]{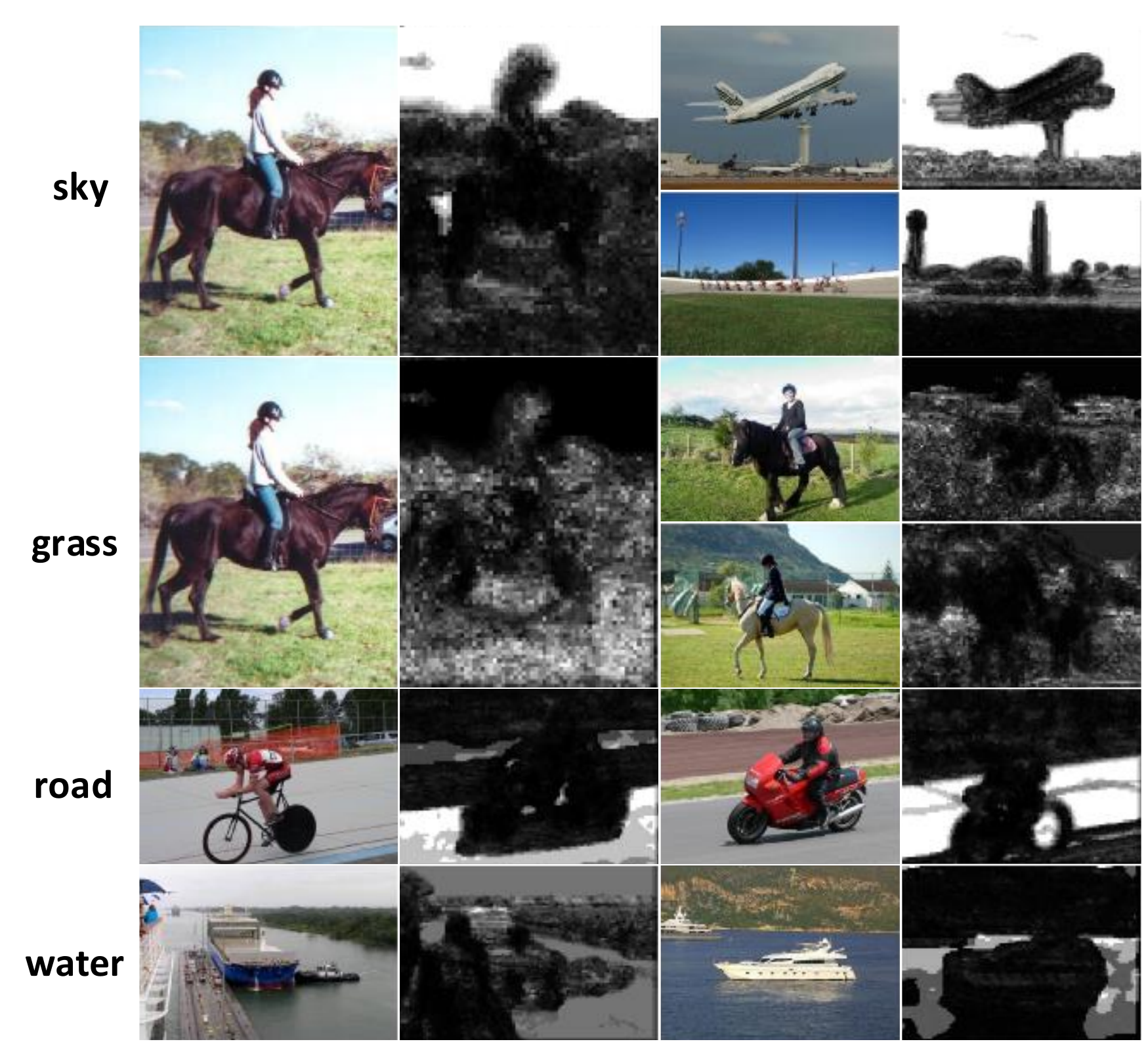}
\end{center}
   \caption{Illustration of the learned background topics. 
   }
\label{fig:backgroundtopic}
\vspace{-10pt}
\end{figure}

\noindent \textbf{Learned background topics } A key ability of our framework is the explicit modelling of background non-annotated data. This allows such irrelevant pixels to be explained, reducing confusion with foreground objects and hence improving localisation accuracy. This is illustrated in Fig.~\ref{fig:backgroundtopic} via plots of the background topic response (heat map). It shows that some of the  background topics have clear semantic meaning, corresponding to common components such as sky, grass, road and water, despite none of these has ever been annotated. Some background components are mixed, e.g. the water topic gives strong response to both water and sky.
But this is understandable because in that image, water and sky are almost visually indistinguishable. 

\noindent \textbf{Further evaluations of our Model }
Table \ref{tab:results_singleview} shows how the performance of our model is reduced without each of several key features, thus validating the usefulness of each of them. 
Specifically, independent learning (\textit{IL}) is an analogue of the  strategy of independent learning of each class used in existing approaches. We train, e.g., 12 models independently for \textit{VOC07-6$\times2$}, each with only one foreground topic. These models are trained on all the data, but all instances without the target object are negative instances.  Table~\ref{tab:results_singleview} verifies that this is sub-optimal  compared to joint learning: each object model no longer has the benefit of utilising the other object models to explain away other foreground objects in multi-label images, thus leading to more confusion within each image.
Without spatially aware representation (\textit{NoSpatial}): The Gaussian representation of appearance within each image enforces spatial compactness, and hence helps to disambiguate object appearance from background appearance. Without learning spatial extent, background patches of similar appearance to objects in the feature space cannot be properly disambiguated, leading to poorer learning and reduced localisation accuracy.  Finally performance is also reduced without using topic-down appearance prior $\pi^0$ (\textit{NoPriorOfapp}) because the model is less likely to converge to a useful local minimal. 

\begin{table}[ht]
\footnotesize
\begin{center}
\begin{tabular}{l | l | c | c }
\hline

\multicolumn{2}{c|}{Method}  & \textit{VOC07-6$\times$2} & \textit{VOC07-20}\\
\hline
\hline
\multirow{3}{50 pt}{Our-Sampling stripped-down} & \textit{IL}  & 42.5    &   29.8   \\
\hhline{~---}
& \textit{NoSpatial}  & 44.6   &   32.1   \\
\hhline{~---}
& \textit{NoPriorOfapp} & 41.7    &  30.4    \\
\hline
\hline

\multirow{2}{50 pt}{Alternative joint learning} & \textit{MIML}  \cite{Zhou07multi-instancemultilabel} & 33.8   &  23.6    \\
\hhline{~---}
& \textit{CorrLDA}   \cite{blei2003annotated_model}    &  34.3   & 27.2   \\
\hline
\hline

\multicolumn{2}{c|}{Our-Sampling  }&  \textbf{ 50.8 }  &  \textbf{ 34.1  }\\
\hline
\end{tabular}
\end{center}
\caption{Further evaluations of individual features of our model and comparing with alternative joint learning approaches.}
\label{tab:results_singleview}
\vspace{-10pt}
\end{table}

\noindent \textbf{Alternative joint learning approaches }
In this experiment we compare other joint multi-instance/weakly-supervised multi-label learning methods, and show that none are effective for WSOL. 
One alternative joint learning approach is to cast WSOL as a MIML learning problem \cite{Zhou07multi-instancemultilabel,XiangyangXue:2011:CMM:2355573.2356568,zha2008mlmi}. Most existing MIML work considers classification. We utilise the model in  \cite{Zhou07multi-instancemultilabel}  and reformulate it for localisation. Specifically, we follow \cite{Deselaers2012} to use the what-is-object boxes to generate bags for each image before applying MIML for localisation. Table \ref{tab:results_singleview} shows that the MIML method under performs, due to the harder discrete optimisation. This, together with the lack the benefit of Bayesian integration of prior knowledge in our model, explains its much poorer result. We also compare with CorrLDA, which was designed for image annotation \cite{blei2003annotated_model}. However its performance is much weaker because it lacks an explicit spatial model and only admits indirect supervision of topics.

\subsection{Computational cost}
\label{cost}
Our model is efficient both in learning and inference, with complexity $\mathcal{O}(NMK)$ for $N$ images, $M$ observations per image, and $K$ classes. The experiments were done on a 2.6Ghz PC with a single-threaded Matlab implementation. Training on all 5,011 VOC07 images required 3 hours and a peak of 6 GB of memory to learn a joint model for 20 classes.
Our Bayesian topic inference process not only enables prior knowledge to be used, but also achieves 10-fold improvements in convergence time compared to EM inference used by most conventional topic models with point-estimated Dirichlet topics. Online inference of a new test image took about 0.5 seconds. For object localisation in training images, direct Gaussian localisation is effectively free and  heat-map sampling took around 0.6 seconds per image. These statistics compare favourably to alternatives: \cite{Deselaers2012}  reports 2 hours to train 100 images; while our Matlab implementations of  \cite{confeccvSivaRX12},  \cite{Sivaiccv2011} and \cite{blei2003annotated_model} took 10, 15 and 20 hours respectively to localise objects for all 5,011 images.

\section{Conclusion}
We presented an effective and efficient model for weakly-supervised object localisation. Our approach surpasses the performance of all prior methods, obtaining state-of-the-art results due to three novel features: joint multi-label learning, a Bayesian formulation, and an explicit spatial model of object location. In addition the computational complexity is favourable compared to prior approaches.
Uniquely with our approach, it is also possible to perform semi-supervised learning and obtain an effective localiser with only a fraction of the annotated training data required by other methods. Moreover, the unlabelled data need not even be sanitised for relevance to the target classes.  In this study we only used simple top-down cues via our Bayesian priors; however this formulation has great potential to enable more scalable learning through cross-class and cross-domain transfer via priors \cite{zhiyuan12,Guillaumin_cvpr12,Kuettel2012}. These contributions bring us significantly closer to the goal of scalable learning of  strong models from weakly-annotated non-purpose collected data on the Internet.

{\footnotesize
\bibliographystyle{ieee}
\bibliography{egbib_zhiyuan}

\begin{thebibliography}{10}\itemsep=-1pt

\bibitem{Alexewhatisobject}
B.~Alexe, T.~Deselaers, and V.~Ferrari.
\newblock What is an object?
\newblock In {\em CVPR}, 2010.

\bibitem{Andrews03supportvector}
S.~Andrews, I.~Tsochantaridis, and T.~Hofmann.
\newblock Support vector machines for multiple-instance learning.
\newblock In {\em NIPS}, 2003.

\bibitem{bishop2006prml}
C.~M. Bishop.
\newblock {\em Pattern Recognition and Machine Learning}.
\newblock Springer, 2006.

\bibitem{blei2003annotated_model}
D.~M. Blei and M.~I. Jordan.
\newblock Modeling annotated data.
\newblock In {\em SIGIR}, 2003.

\bibitem{BleiLDA2003}
D.~M. Blei, A.~Y. Ng, and M.~I. Jordan.
\newblock Latent dirichlet allocation.
\newblock {\em JMLR}, 2003.

\bibitem{CabralDCB11}
R.~S. Cabral, F.~{De la Torre}, J.~P. Costeira, and A.~Bernardino.
\newblock Matrix completion for multi-label image classification.
\newblock In {\em NIPS}, 2011.

\bibitem{CaoFei-Fei2007}
L.~Cao and L.~Fei-Fei.
\newblock Spatially coherent latent topic model for concurrent object
  segmentation and classification.
\newblock In {\em ICCV}, 2007.

\bibitem{Crandalleccv06}
D.~Crandall and D.~Huttenlocher.
\newblock Weakly supervised learning of part-based spatial models for visual
  object recognition.
\newblock In {\em ECCV}, 2006.

\bibitem{DengECCV2010}
J.~Deng, A.~C. Berg, K.~Li, and L.~Fei-Fei.
\newblock What does classifying more than 10,000 image categories tell us?
\newblock In {\em ECCV}, 2010.

\bibitem{Deselaers2012}
T.~Deselaers, B.~Alexe, and V.~Ferrari.
\newblock Weakly supervised localization and learning with generic knowledge.
\newblock {\em IJCV}, 2012.

\bibitem{pascalvoc2007}
M.~Everingham, L.~Van~Gool, C.~K.~I. Williams, J.~Winn, and A.~Zisserman.
\newblock The {PASCAL} {V}isual {O}bject {C}lasses {C}hallenge 2007 {(VOC2007)}
  {R}esults.
\newblock
  http://www.pascal-network.org/challenges/VOC/voc2007/workshop/index.html.

\bibitem{feifei2006one_shot}
L.~Fei-Fei, R.~Fergus, and P.~Perona.
\newblock One-shot learning of object categories.
\newblock {\em TPAMI}, 2006.

\bibitem{Felzenszwalb2012partbased}
P.~Felzenszwalb, R.~Girshick, D.~McAllester, and D.~Ramanan.
\newblock Object detection with discriminatively trained part-based models.
\newblock {\em TPAMI}, 2010.

\bibitem{fu2012attribsocial}
Y.~Fu, T.~Hospedales, T.~Xiang, and S.~Gong.
\newblock Attribute learning for understanding unstructured social activity.
\newblock In {\em ECCV}, 2012.

\bibitem{houghforest2012}
J.~Gall, A.~Yao, N.~Razavi, L.~Van~Gool, and V.~Lempitsky.
\newblock Hough forests for object detection, tracking, and action recognition.
\newblock {\em TPAMI}, 2011.

\bibitem{Guillaumin_cvpr12}
M.~Guillaumin and V.~Ferrari.
\newblock { Large-scale Knowledge Transfer for Object Localization in
  ImageNet}.
\newblock In {\em CVPR}, 2012.

\bibitem{tim2011tpami}
T.~Hospedales, J.~Li, S.~Gong, and T.~Xiang.
\newblock Identifying rare and subtle behaviors: A weakly supervised joint
  topic model.
\newblock {\em TPAMI}, 2011.

\bibitem{Kuettel2012}
D.~Kuettel, M.~Guillaumin, and V.~Ferrari.
\newblock Segmentation propagation in imagenet.
\newblock In {\em ECCV}, 2012.

\bibitem{LempitskICCV2009}
V.~Lempitsky, P.~Kohli, C.~Rother, and T.~Sharp.
\newblock Image segmentation with a bounding box.
\newblock In {\em ICCV}, 2009.

\bibitem{LiSocherFeiFei2009}
L.-J. Li, R.~Socher, and L.~Fei-Fei.
\newblock Towards total scene understanding:classification, annotation and
  segmentation in an automatic framework.
\newblock In {\em CVPR}, 2009.

\bibitem{Maron98aframework}
O.~Maron and T.~Lozano-Pérez.
\newblock A framework for multiple-instance learning.
\newblock In {\em NIPS}, 1998.

\bibitem{Nguyenweakly2011}
M.~Nguyen, L.~Torresani, F.~de~la Torre, and C.~Rother.
\newblock Weakly supervised discriminative localization and classification: a
  joint learning process.
\newblock In {\em ICCV}, 2009.

\bibitem{nguyen2010svm_miml}
N.~Nguyen.
\newblock A new svm approach to multi-instance multi-label learning.
\newblock In {\em ICDM}, pages 384--392, 2010.

\bibitem{Pandeyiccv2011}
M.~Pandey and S.~Lazebnik.
\newblock Scene recognition and weakly supervised object localization with
  deformable part-based models.
\newblock {\em ICCV}, 2011.

\bibitem{Philbinijcv2010}
J.~Philbin, J.~Sivic, and A.~Zisserman.
\newblock Geometric latent dirichlet allocation on a matching graph for
  large-scale image datasets.
\newblock {\em IJCV}, 2011.

\bibitem{zhiyuan12}
Z.~Shi, P.~Siva, and T.~Xiang.
\newblock Transfer learning by ranking for weakly supervised object annotation.
\newblock In {\em BMVC}, 2012.

\bibitem{confeccvSivaRX12}
P.~Siva, C.~Russell, and T.~Xiang.
\newblock In defence of negative mining for annotating weakly labelled data.
\newblock In {\em ECCV}, 2012.

\bibitem{Sivaiccv2011}
P.~Siva and T.~Xiang.
\newblock Weakly supervised object detector learning with model drift
  detection.
\newblock In {\em ICCV}, 2011.

\bibitem{Sivic05b}
J.~Sivic, B.~C. Russell, A.~A. Efros, A.~Zisserman, and W.~T. Freeman.
\newblock Discovering object categories in image collections.
\newblock In {\em ICCV}, 2005.

\bibitem{sudderth2008tdp_visual}
E.~B. Sudderth, A.~Torralba, W.~T. Freeman, and A.~S. Willsky.
\newblock Describing visual scenes using transformed objects and parts.
\newblock {\em IJCV}, 2008.

\bibitem{wangbleifeifei08}
C.~Wang, D.~Blei, and L.~Fei-Fei.
\newblock Simultaneous image classification and annotation.
\newblock In {\em CVPR}, 2009.

\bibitem{winn2004vmp}
J.~Winn and C.~M. Bishop.
\newblock Variational message passing.
\newblock {\em JMLR}, 2005.

\bibitem{XiangyangXue:2011:CMM:2355573.2356568}
X.~Xue, W.~Zhang, J.~Zhang, B.~Wu, J.~Fan, and Y.~Lu.
\newblock Correlative multi-label multi-instance image annotation.
\newblock In {\em ICCV}, 2011.

\bibitem{zha2008mlmi}
Z.-J. Zha, X.-S. Hua, T.~Mei, J.~Wang, G.-J. Qi, and Z.~Wang.
\newblock Joint multi-label multi-instance learning for image classification.
\newblock In {\em CVPR}, 2008.

\bibitem{Zhou07multi-instancemultilabel}
Z.~Zhou and M.~Zhang.
\newblock Multi-instance multilabel learning with application to scene
  classification.
\newblock In {\em NIPS}, 2007.

\bibitem{bMCL2012}
J.-Y. Zhu, J.~Wu, Y.~Wei, E.~I.-C. Chang, and Z.~Tu.
\newblock Unsupervised object class discovery via saliency-guided multiple
  class learning.
\newblock In {\em CVPR}, 2012.

\bibitem{zhu2007sslsurvey}
X.~Zhu.
\newblock Semi-supervised learning literature survey.
\newblock Technical Report 1530, University of Wisconsin-Madison Department of
  Computer Science, 2007.

\end{thebibliography}
}

\end{document}